\documentclass[10pt,twocolumn,letterpaper]{article}

\usepackage{cvpr}
\usepackage{times}
\usepackage{epsfig}
\usepackage{graphicx}
\usepackage{amsmath}
\usepackage{amssymb}
\usepackage{natbib}
\usepackage{multirow}
\usepackage[breaklinks=true,bookmarks=false]{hyperref}

\cvprfinalcopy 

\def\cvprPaperID{****} 

\begin{document}

\title{Recurrent convolutional neural network for the surrogate modeling of subsurface flow simulation}

\author{Hyung Jun Yang,
\and Timothy Yeo, 
\and Jaewoo An\thanks{Non-CS231N contributor} \\
Stanford University\\
Department of Energy Resources Engineering, Stanford University, Stanford, CA, USA,\\
{\tt\small hjyang3@stanford.edu
           timyeo@stanford.edu
           jaewooan@stanford.edu}
}

\maketitle


\section{Abstract}

The quantification of uncertainty on fluid flow in porous media is often hampered by multi-scale heterogeneity and insufficient site characterization. Monte-Carlo simulation (MCS), which runs numerical simulations for a large number of realization of input parameters , becomes infeasible when simulation cost is expensive or the degree of uncertainty is large. Many deep-neural-network-based methods are developed in order to replace the numerical flow simulation, but previous studies focused only on generating several snapshots of outputs at the fixed time steps, and lack to reflect the time dependent property of simulation data. Recently, the convolutional long short term memory (ConvLSTM) is utilized to deal with time series image data. Here, we propose to combine SegNet with ConvLSTM layers for the surrogate modeling of numerical flow simulation. The results show that the proposed method improves the performance of SegNet based surrogate model remarkably when the output of the simulation is time series data. 

\section{Introduction}

The reliable predictions of fluid flow in subsurface environments are compromised by multi-scale heterogeneity and insufficient site characterization. These factors introduce uncertainty in input parameters (e.g., permeability, porosity, and initial saturation), which renders model outputs (e.g., saturation and production rate at each time steps) uncertain as well. Monte Carlo simulation (MCS) has been widely used to quantify this output uncertainty. It repeatedly performs numerical simulations (i.e., finite volume simulation) for each realization of input parameters until its solution converges. However, this approach requires a large number of realizations, especially when the multiple parameters are uncertain or the uncertainty range is large. When a single model run is computationally expensive, the use of MCS might become unfeasible. 
\\  
To improve the computational efficiency of the previous approaches, numerous data-driven models have emerged. The recent achievements in deep learning and parallel computation have evoked the works aiming to replace the numerical flow simulation with the deep-neural-network-based methods. \cite{raissi2017physics} proposed physics-informed neural networks to solve PDEs without data. The main idea of physics-informed (or physics-constrained) learning is to incorporate physical knowledge through constraint learning, i.e. learning the models by minimizing the violation of the physical constraints as well as conventional data misfit. \cite{Nabian_2019, karumuri2020simulator, sun2020surrogate} used fully connected neural network for physics constrained surrogate modeling. \cite{zhu2019physics, mo2019deep} used convolutional encoder-decoder for physics-constrained learning to create the surrogate models for flow simulation. \cite{yang2019adversarial} used deep generative models for the physics constrained surrogate modeling and propagating the uncertainty. 

While the outputs of the numerical simulations are generally time-series data (e.g., saturation maps at each time step), these previous approaches only focused on generating the several snapshots of outputs at the fixed time steps. Their approaches not only restrict the applicability of surrogate models but fail to reflect the time dependency of simulation outputs. To overcome these shortcomings, we propose recurrent convolutional neural network for the surrogate modeling of flow and transport. Utilizing the recurrent neural network and convolutional neural network (CNN) simultaneously, we expect to improve the performance of our neural network with time series outputs. 

\section{Problem Statement}
\subsection{Multi-phase flow problem}

The multi-phase flow in porous media is generally modeled by solving the mass balance equation defined as \eqref{eq:mass_balance}, where the pressure $p$ and saturation $S$ are considered as independent variables.
\begin{align}
\label{eq:mass_balance}
\frac{\partial(\rho_k \phi S_k)}{\partial t} + \nabla \cdot \left( \rho_k \boldsymbol{u_k} \right) - \rho_k q_k = 0,   k = w, nw.
\end{align}
$\rho$, $\phi$, $q$ are the mass density,  the porosity,  and volumetric source/sink term, respectively. The subscripts w, nw represent wetting and non-wetting phases. The Darcy velocity $\textbf{u}$ can be expressed as follows.
\begin{align}
\label{eq:darcy}
\textbf{u} = -\frac{k k_{rk}}{\mu_k} \left( \nabla p - \rho_k \boldsymbol{g} \right), \quad k = w, nw.
\end{align}
Here, $k, k_r$ are the absolute and relative permeability, $\mu$ is the fluid viscosity, and $\textbf{g}$ is the gravitational acceleration.

Equation~\ref{eq:mass_balance}, together with Equation~\ref{eq:darcy}, becomes highly nonlinear due to the nonlinear characteristics of parameters such as relative permeability $k_r$ (a function of saturation) and heterogeneous permeability field.

\subsection{Surrogate modeling}
Due to the uncertain nature of subsurface environments, the parameters such as permeability $k$ are uncertain in most cases. For reliable uncertainty quantification, MCS based methods repeatedly solve the multi-phase problem numerically. Since this numerical simulation requires a lot of computation, our objective is to construct the surrogate model using the deep neural network. 

In this work, input $X$ is a static permeability image and output $Y$ is the saturation or pressure map at each time steps. The output can be considered as time-series image data (i.e., video-like data). Now, our surrogate modeling problem can be interpreted as image-to-image regression problem with regression function $\eta: X \xrightarrow[]{} Y$. Therefore, we can employ previous techniques pertaining to image-to-image regression problem to solve our problem. 

\subsection{Dataset}
\begin{figure}[t]
\begin{center}
\includegraphics[width=0.5\linewidth]{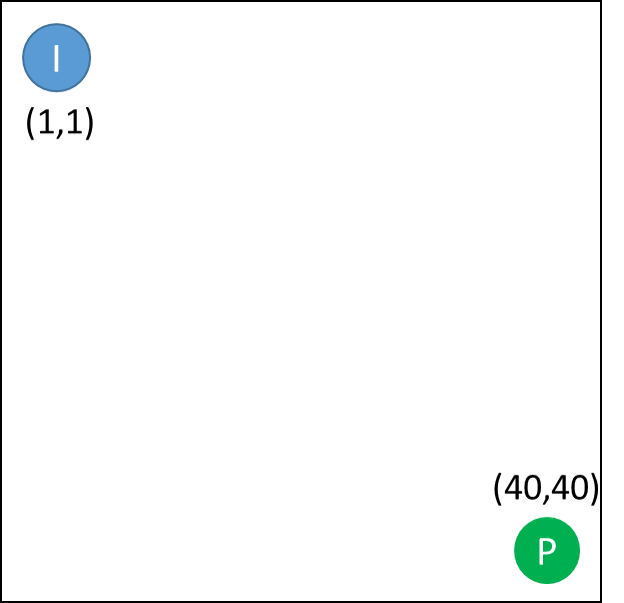}
\end{center}
   \caption{Reservoir configuration (standard condition): 10 $m^3/sec$ of water injection at (1,1) and 10 $m^3/sec$ of liquid production at (40,40)}
\label{fig:Reservoir_config}
\end{figure}

In this paper, a two-dimensional incompressible oil and water flow problem is analyzed with the five-spot well pattern: four injection wells on each corner and one production well at the center. With the assumption of symmetry on reservoir characteristics and well conditions with respect to x and y axes, the total configuration can be simplified to one quarter of the domain which includes one water injection well and one oil production well (figure \ref{fig:Reservoir_config}).

The input of our data is permeability distribution generated by the sequential Gaussian simulation. The corresponding outputs are water saturation and pressure distributions for 30 time steps during 300 days. The data generated by the commercial reservoir simulator ECLIPSE is considered as true data. In this paper, 4,000 sets of data are generated for training. Additional information for dataset is described in Table~\ref{table:simulation_input}. 

\begin{table}
\begin{center}
\begin{tabular}{|l|c||c|c|}
\hline
Parameter & Value &  Parameter & value  \\
\hline\hline
Domain (m) & 40, 40, 1 & Grid size (m) & 1\\
Time (year) & 10 & \# of time steps & 10\\
Porosity & 1 & Top depth (m) & 5,000\\
$\rho_{oil}\ (kg/m^3)$ & 600 & $\rho_{water}\ (kg/m^3)$ & 999\\
$\mu_{oil}\ (cP)$ & 2 & $\mu_{water}\ (cP)$ & 1\\
Input dims. & (40,40) & Output dims & (10,40,40) \\
\hline
\end{tabular}
\end{center}
\caption{Dataset information}
\label{table:simulation_input}
\end{table}
\section{Technical approach}

\subsection{Physics informed learning}
We propose to use physics-informed deep neural network to construct a reliable prediction tool with only scarce observed data. The physics-informed machine learning is recently developed and utilized by \cite{raissi2017physics, chang2017physics, yang2019adversarial, sun2020surrogate}. There are some variants, but most of the methods impose the physical constraints by adding residual terms to the objective function. This residual term measures the violation of physical laws. As a result, our training minimizes not only the violation of physical laws but also the discrepancy with the observed data. We define the new loss function as 
\begin{equation}\label{eq:newloss}
 L_{total} = L_{data} + \lambda L_{physics} 
 \end{equation}
 \begin{equation}
 L_{data} = \frac{1}{m} \sum_{i=0}^{m} [y_{pred}^{(i)} - y_{true}^{(i)}][y_{pred}^{(i)} - y_{true}^{(i)}]^T \end{equation}
\begin{equation}L_{physics} = \int_\Omega (\frac{\partial(\rho_k \phi S_k)}{\partial t} + \nabla \cdot \left( \rho_k \boldsymbol{u_k} \right) - \rho_k q_k) dV
\end{equation}
where $\Omega$ is the spatial domain, and $\lambda$ is the constant that scales the physical loss. 
We can see that $L_{data}$ is mean squared error between true data and predicted data. $L_{physics}$ which measures the violation of physics law defined in Equation~\eqref{eq:mass_balance}.

\subsection{Convolutional Long Short Term Memory }
LSTM (Long Short Term Memory) as a special recurrent neural network (RNN) structure has proven to be useful for general-purpose sequence modeling (\cite{LSTM_1997}, \cite{Graves_2013}). Through the memory cell and various gates, LSTM solves the problem of the vanishing gradient which is the major problem of the vanila RNN. FC-LSTM (Fully Connected LSTM) can be seen as a multivariate version of LSTM where the input, cell output and states are all 1D vectors (\cite{ConvLSTM}). 
The major drawback of FC-LSTM is that it uses full connection both in input-to-state, and state-to-state transitions and therefore, no spatial information is encoded. In order to better handle the spatio-temproal data, Convolutional LSTM or ConvLSTM was first proposed by \cite{ConvLSTM}.  The key equations of ConvLSTM can be expressed as follows.
\begin{equation}
\begin{aligned}
\label{eq-convLSTM}
i_{t} = \sigma\left( W_{xi} * x_t + W_{hi}*h_{t-1} + W_{ci} \circ c_{t-1} + b_i   \right) \\
f_{t} = \sigma\left( W_{xf} * x_t + W_{hf} * h_{t-1} + W_{cf} \circ c_{t-1} + b_f \right) \\
c_{t} = f_{t} \circ c_{t-1} + i_t \circ \tanh(W_{wc}*x_t + W_{hc} * h_{t-1} + b_c)  \\
o_{t} = \sigma \left( W_{xo}*x_t + W_{ho} * h_{t-1} + W_{co} \circ c_t +b_o \right) \\
h_t = o_t \circ \tanh(c_t)
\end{aligned}
\end{equation}

where, '$*$' denotes the convolution operator and '$\circ$' denotes the Hadamard product. ConvLSTM is extended from the FC-LSTM by changing full connection in input-to-state and state-to-state transitions to convolution operations and therefore, the spatial information is encoded. Note that all inputs, state variables, memory cells, and gates are the 3D tensors where the last two dimensions are spatial dimensions unlike the FC-LSTM.

\subsection{Convolutional encoder-decoder model}
\begin{figure}[t]
\begin{center}
   \includegraphics[width=1.0\linewidth]{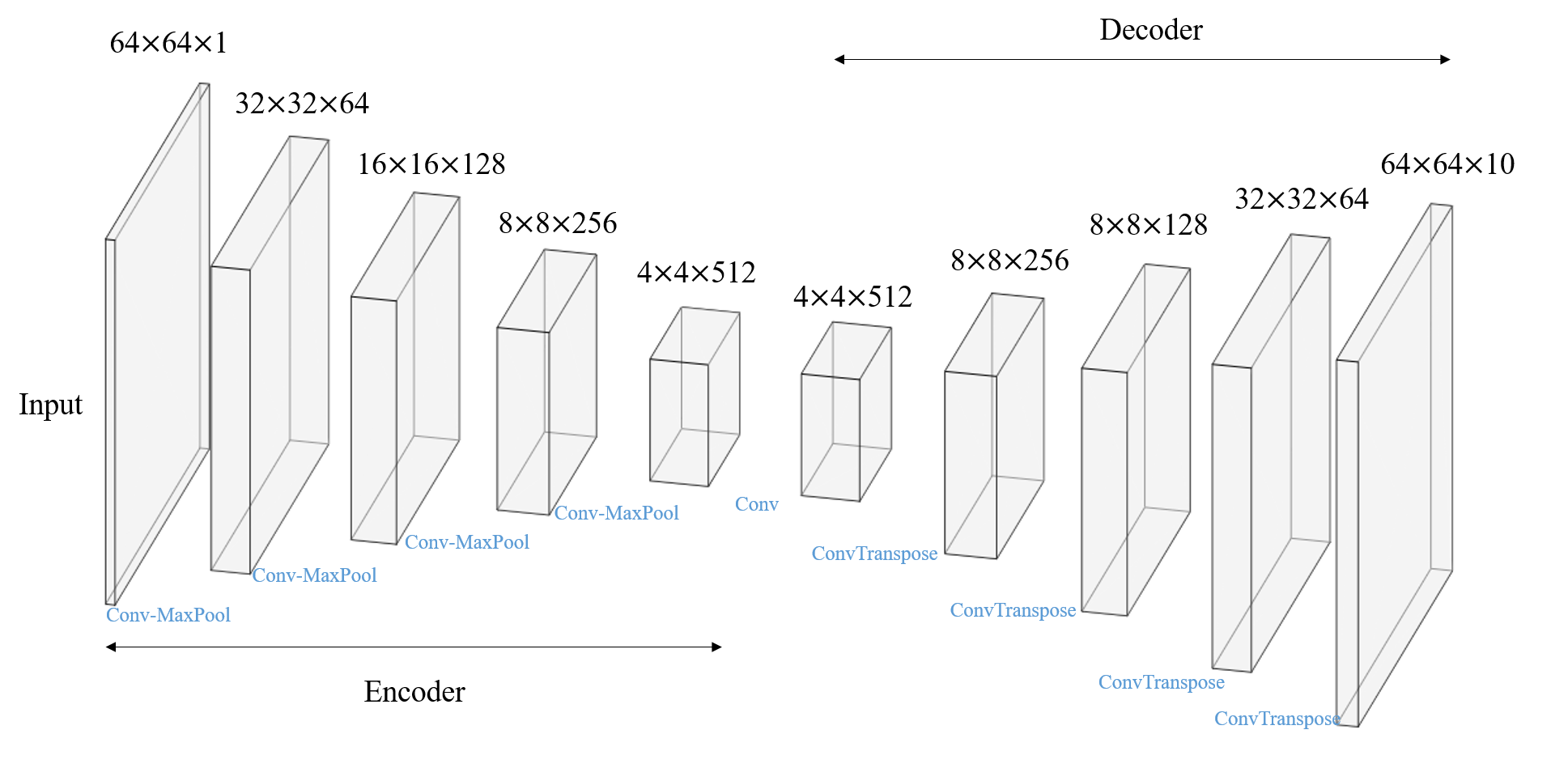}
\end{center}
   \caption{Architecture of SegNet}
\label{fig:SegNet}
\end{figure}

As we discussed, our surrogate model can be transformed to image-to-image regression problem. For solving image-to-image regression problem (e.g., image restoration, pixel-wise segmentation), convolutional encoder-decoder has been widely used. The encoder in the network computes progressively coarse-scale abstract features as the receptive fields increase with the depth of the encoder. During encoding process, the spatial resolution of the feature maps is reduced via a down-sampling operation (e.g., max pooling) while the decoder computes feature maps of increasing resolution via unpooling or transpose convolution operation. Through this encoder-decoder structure, the network can model local and global features of images. Among many variations of the convolutional encoder-decoder network, SegNet proposed by \cite{badrinarayanan2017segnet} is one of the most popular methods. It employs the first 13 convolutional layers of VGG-16 as its encoder and each encoder layer has a corresponding decoder layer. Figure~\ref{fig:SegNet} shows the general structure of SegNet that we used in this paper. 

\subsection{Proposed method: SegNet-ConvLSTM}
\begin{figure}[t]
\begin{center}
   \includegraphics[width=1.0\linewidth]{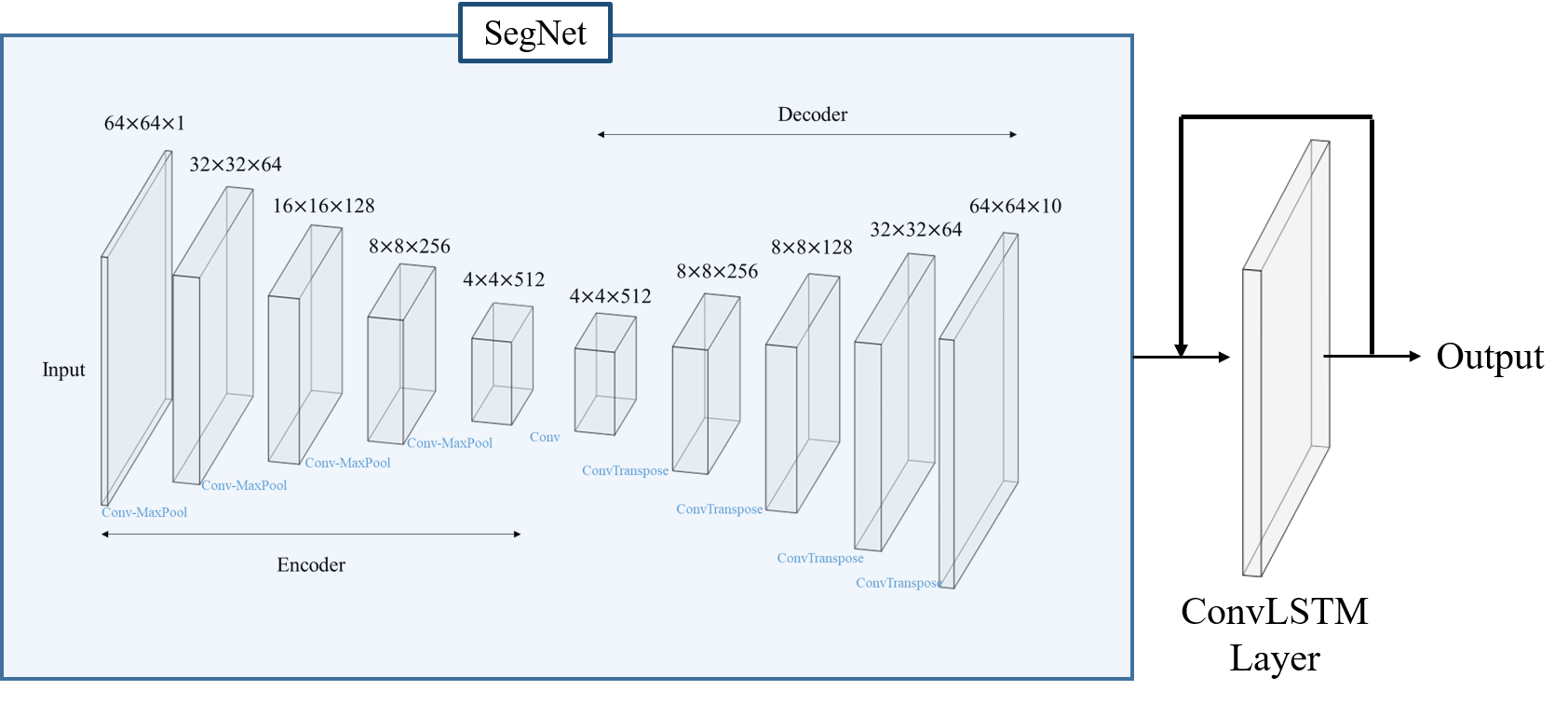}
\end{center}
   \caption{Architecture of SegNet-ConvLSTM}
\label{fig:ConvLSTM}
\end{figure}

As we discussed in the previous sections, our output data is time-series image data (i.e., video-like data). To utilize this property and improve the performance of SegNet, we propose SegNet-ConvLSTM. Figure~\ref{fig:ConvLSTM} shows the general architecture of SegNet-LSTM. The only difference between SegNet and SegNet-ConvLSTM is that one convolutional LSTM layer is added next to the last layer of SegNet. Using ConvLSTM at the last step, we expect to predict the time-dependent output with better accuracy. Since we added only one layer, the small number (\~{} 100) of parameters is added to the original SegNet. Therefore, the increases of computation and memory requirements compared to SegNet are also relatively small. 

\section{Results}
\subsection{Case 1: surrogate model without physical constraint}
In this case, we only use $L_{data}$ in \eqref{eq:newloss} as our loss function. It means that we do not use any physics-informed learning. We compare our proposed SegNEt-ConvLSTM with plain Segnet to illustrate the performance of the proposed method. To get the intermediate results with low computational cost, we currently employ a simplified version of SegNet having only 7 convolutional layers as its encoder. The overall structure of simplified SegNet is the same as the original version of SegNet. Table~\ref{tab:Parameters} summarizes the parameters related to experiments and hyper-parameters we used for networks. All parameter values are optimized based on the grid search results. 

\begin{figure}[t]
\begin{center}
   \includegraphics[width=1.0\linewidth]{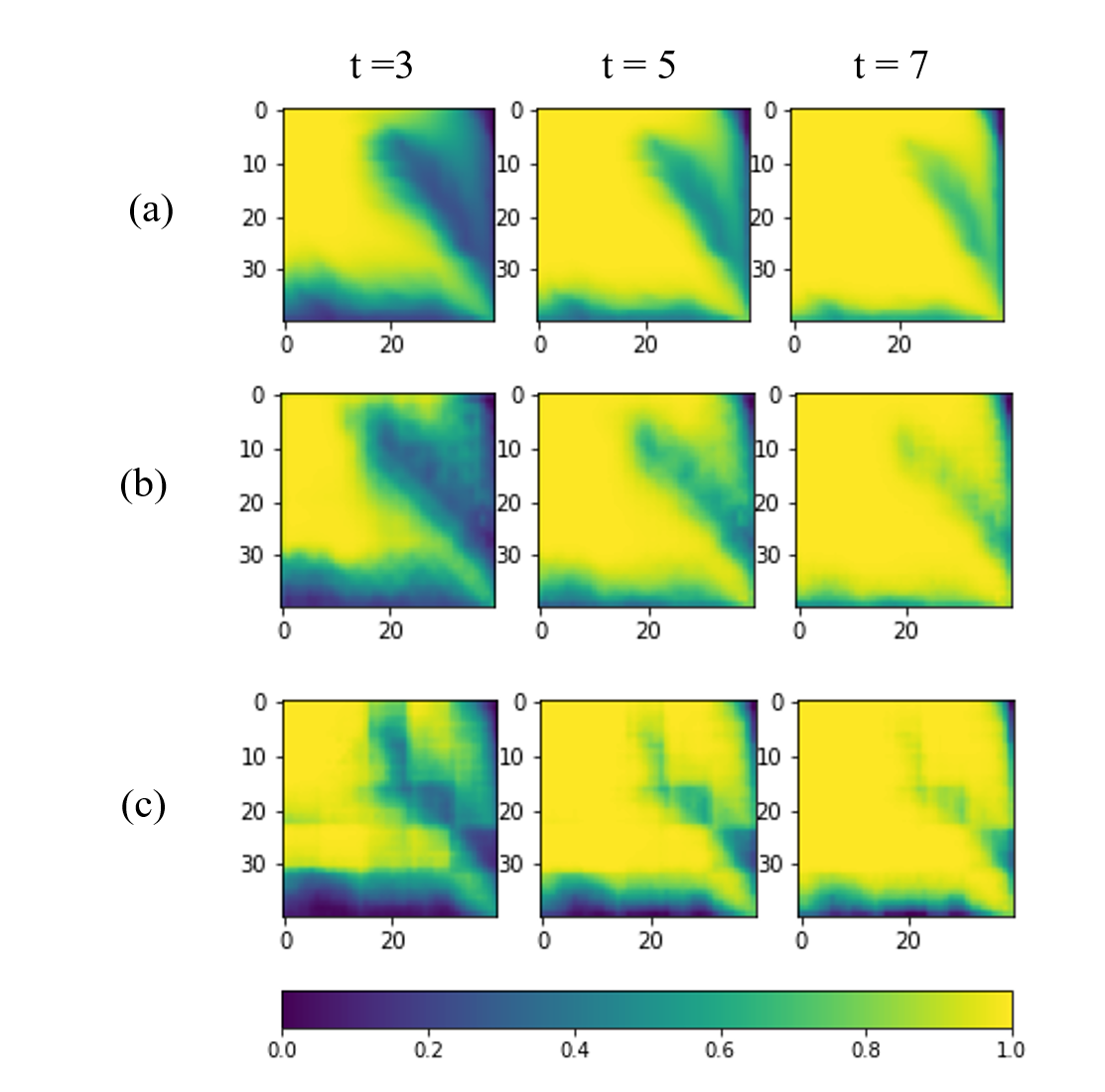}
\end{center}
   \caption{Case1: Comparison of saturation predictions at three different time steps. (a) numerical simulation, (b) SegNet, and (c) SegNet-LSTM.}
\label{fig:long}
\label{fig:SamplePrediction}
\end{figure}

Table~\ref{tab:MSE_loss} compares the training, validation, and test loss obtained by SegNet and SegNet-ConvLSTM. SegNet-ConvLSTM which is designed to deal with time-series data provides lower MSE (mean squared error) loss compare to SegNet in all datasets. Figure~\ref{fig:SamplePrediction} shows that SegNet-ConvLSTM predicts the evolution of saturation profiles with good accuracy, while the plain SegNet shows a large disagreement with true data. The results demonstrate that combining convolutional LSTM layer to SegNet improves the performance of SegNet when the output is time-series image data. 

\begin{table}

\begin{center}
\begin{tabular}{|l|c|c|}
\hline
Parameters & Values  \\
\hline\hline
No. of training data & 3,550 \\
No. of validation data & 300 \\
No. of test data & 150 \\
No. of time steps & 10 \\
Input dim. & (40,40,1) \\ 
Output dim. & (40,40,10,1) \\ 
Learning rate & 3e-2 \\
Dropout rate & 0.1 \\
Regularization & l2-norm \\
Weight decay of regularization & 1e-4 \\
No. of epochs & 150 \\
\hline
\end{tabular}
\end{center}
\caption{Parameters used for Case 1}
\label{tab:Parameters}
\end{table}

\begin{table}
\begin{center}
\begin{tabular}{|l|c|c|c|}
\hline
Method & SegNet &  SegNet-ConvLSTM  \\
\hline\hline
Training loss & 6.88e-4  & 4.01e-4 \\
Validation loss & 3.70e-3 & 2.99e-3 \\
Test loss & 3.75e-3 & 3.04e-3\\
\hline
\end{tabular}
\end{center}
\caption{Mean squared error loss for Case 1}
\label{tab:MSE_loss}
\end{table}

\subsection{Case 2: surrogate model with physical constraint}

 The results from case 1 show that we can get reasonably good predictions by minimizing only data loss. However, our surrogate model should not violate the physical laws (e.g., the conservation law for mass or energy) for practical implementation. In case 2, we use new loss function \eqref{eq:newloss} that consider both data and physical losses for training the neural network. The architecture of neural network and hyper-parameters for case 2 is same as in case 1 except for the definition of loss function. The details of dataset used for case 2 is summarized in table~\ref{tab:Case2Parameters}. 
  
\begin{figure}[t]
\begin{center}
   \includegraphics[width=1.0\linewidth]{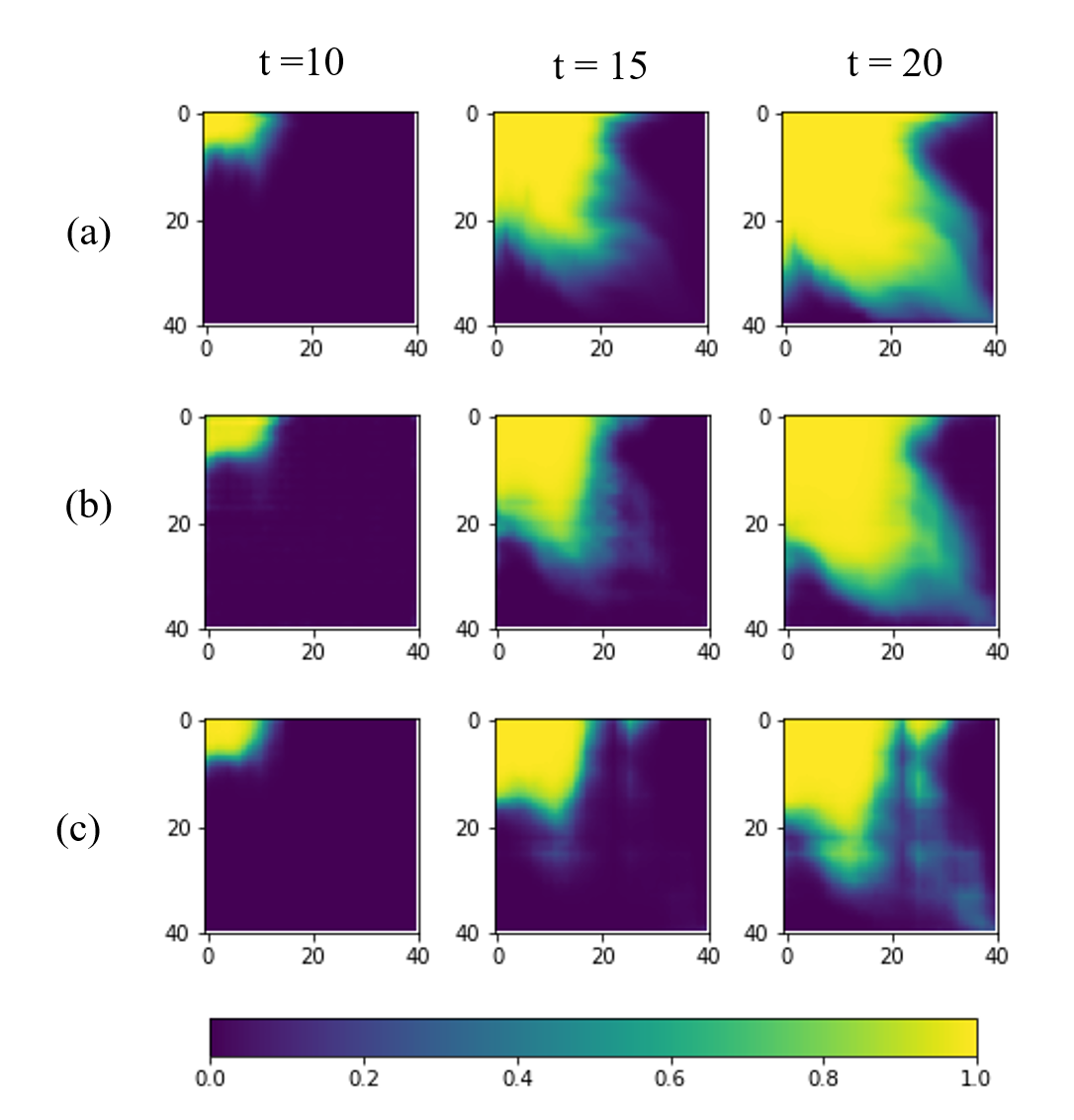}
\end{center}
   \caption{Case2: Comparison of saturation predictions at three different time steps. (a) numerical simulation, (b) SegNet, and (c) SegNet-LSTM.}
\label{fig:long}
\label{fig:Case2Prediction}
\end{figure}

 Table~\ref{tab:Case2_loss} compares the physical loss and data misfit obtained by SegNet and SegNet-ConvLSTM. Similar to Case 1, SegNet-ConvLSTM provides lower MSE (mean squared error) and physical loss compare to SegNet in all datasets. Figure~\ref{fig:Case2Prediction} shows the prediction results for one input test sample. The visual inspection of ~\ref{fig:Case2Prediction} verifies that the prediction from SegNet-ConvLSTM is closer to the true data compare to plain SegNet. It is shown that the SegNet does not well capture the propagation of saturation with time and produces physically unreasonable artifacts to image. 
  
\begin{figure}[t]
\begin{center}
   \includegraphics[width=1.0\linewidth]{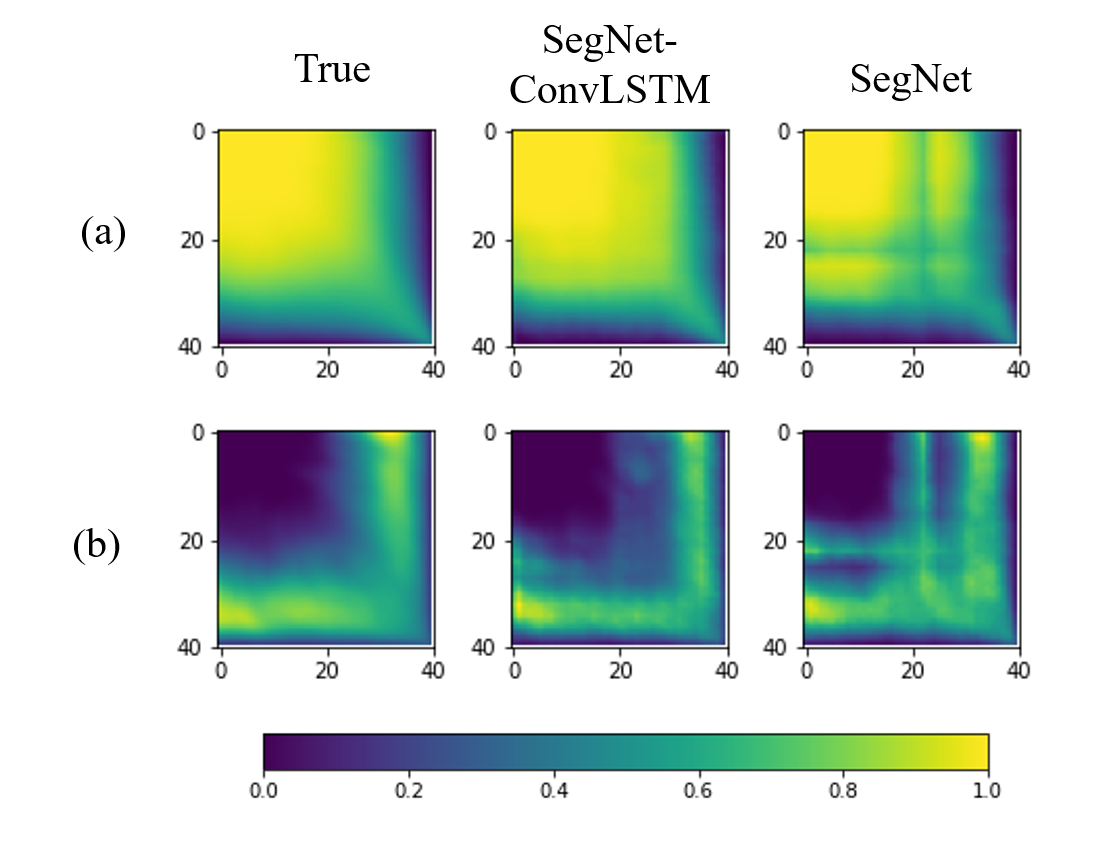}
\end{center}
   \caption{Case2: Comparison of predicted statistical parameters (mean and variance) at $t=15$. (a) numerical simulation, (b) SegNet, and (c) SegNet-LSTM.}
\label{fig:long}
\label{fig:Case2stats}
\end{figure}

  To verify the performance of the surrogate models on uncertainty quantification, we computed two statistical parameters (i.e., mean and varaince) from 1,000 test results and compared with numerical simulation results which are regarded as true data. The visual inspection of Figure~\ref{fig:Case2stats} shows that the proposed mean and variance computed with SegNet-ConvLSTM are in close agreement with true data. Yet, SegNet fails to predict the statistical moments with reasonable accuracy. These results indicate that the proposed SegNet-ConvLSTM improves the performance of CNN based surrogate models significantly.

  \begin{table}

\begin{center}
\begin{tabular}{|l|c|c|}
\hline
Parameters & Values  \\
\hline\hline
No. of training data & 1,550 \\
No. of validation data & 150 \\
No. of test data & 150 \\
No. of time steps & 50 \\
Input dim. & (40,40,1) \\ 
Output dim. & (40,40,50,1) \\ 
Scale parameter for physical loss($\lambda$) & 0.3 \\ 
\hline
\end{tabular}
\end{center}
\caption{Details of dataset used for case 2}
\label{tab:Case2Parameters}
\end{table}

\begin{table}
\begin{center}
\begin{tabular}{|l|c|c|c|}
\hline
Loss & Method & SegNet &  SegNet-ConvLSTM  \\
\hline\hline
\multirow{3}{*}{MSE} & Training Set & 5.21e-3  & 2.3e-3 \\
& Validation loss & 3.70e-2 & 1.12e-2 \\
& Test loss & 3.55e-2 & 1.23e-2\\
\hline
\multirow{3}{*}{Physical} & Training Set & 5.13e-2  & 4.01e-2 \\
& Validation loss & 9.73e-2 & 8.79e-2 \\
& Test loss & 9.28e-2 & 9.14e-2\\

\hline
\end{tabular}
\end{center}
\caption{Mean squared error loss for Case 2}
\label{tab:Case2_loss}
\end{table}

\section{Conclusion $\&$ Discussion}
The reliable surrogate modeling of flow and transport in subsurface environment has been considered as notoriously hard due to the the high non-linearity and uncertainty.  In our work, we propose to use convolutional-LSTM-based deep neural network to enhance the performance of deep-learning-based surrogate modeling. We added only one convolutional-LSTM layer to the conventional SegNet-like convolutional encoder-decoder model. This increases relatively few parameters and slight computational cost, but enables to deal with time series data.

We applied our proposed method for the surrogate modeling of multi-phase flow in porous media . The first case minimized only data misfit and the second one minimized both data misfit and the violation of physical law. In both cases, the accuracy measures show that the proposed SegNet-ConvLSTM combining SegNet and convolutional LSTM outperforms the plain SegNet. The statistical parameters from proposed method and numerical simulation results are also in good agreement. Thus, it is proven that SegNet-ConvLSTM quantifies the uncertainty on flow response reliably. 
In the future steps, extending toward more complex flow simulations such as 3D3P (3 Dimensional 3 Phase) flow simulation could be a good target.

\section{Contribution $\&$ Acknowledgement}
We acknowledge the Department of Energy Resources Engineering at Stanford University for the Eclipse software license. The experiments are conducted on the cluster of Stanford
Center for Computational Earth \& Environmental Sciences (CEES). The provided computational resources are greatly appreciated.

{\small
\bibliography{egbib}

\begin{thebibliography}{12}
\providecommand{\natexlab}[1]{#1}
\providecommand{\url}[1]{\texttt{#1}}
\expandafter\ifx\csname urlstyle\endcsname\relax
  \providecommand{\doi}[1]{doi: #1}\else
  \providecommand{\doi}{doi: \begingroup \urlstyle{rm}\Url}\fi

\bibitem[Badrinarayanan et~al.(2017)Badrinarayanan, Kendall, and
  Cipolla]{badrinarayanan2017segnet}
Vijay Badrinarayanan, Alex Kendall, and Roberto Cipolla.
\newblock Segnet: A deep convolutional encoder-decoder architecture for image
  segmentation.
\newblock \emph{IEEE transactions on pattern analysis and machine
  intelligence}, 39\penalty0 (12):\penalty0 2481--2495, 2017.

\bibitem[Chang et~al.(2017)Chang, Dinh, and Cetiner]{chang2017physics}
Chih-Wei Chang, Nam Dinh, and Sacit~M Cetiner.
\newblock Physics-constrained machine learning for two-phase flow simulation
  using deep learning-based closure relation.
\newblock In \emph{American Nuclear Society Winter Meeting, Washington, DC},
  pages 1749--1752, 2017.

\bibitem[Graves(2013)]{Graves_2013}
Alex Graves.
\newblock Generating sequences with recurrent neural networks.
\newblock \emph{arXiv preprint arXiv:1308.0850}, 2013.

\bibitem[Hochreiter and Schmidhuber(1997)]{LSTM_1997}
Sepp Hochreiter and Jürgen Schmidhuber.
\newblock Long short-term memory.
\newblock \emph{Neural computation}, 9\penalty0 (8):\penalty0 1735--1780, 1997.

\bibitem[Karumuri et~al.(2020)Karumuri, Tripathy, Bilionis, and
  Panchal]{karumuri2020simulator}
Sharmila Karumuri, Rohit Tripathy, Ilias Bilionis, and Jitesh Panchal.
\newblock Simulator-free solution of high-dimensional stochastic elliptic
  partial differential equations using deep neural networks.
\newblock \emph{Journal of Computational Physics}, 404:\penalty0 109120, 2020.

\bibitem[Mo et~al.(2019)Mo, Zhu, Zabaras, Shi, and Wu]{mo2019deep}
Shaoxing Mo, Yinhao Zhu, Nicholas Zabaras, Xiaoqing Shi, and Jichun Wu.
\newblock Deep convolutional encoder-decoder networks for uncertainty
  quantification of dynamic multiphase flow in heterogeneous media.
\newblock \emph{Water Resources Research}, 55\penalty0 (1):\penalty0 703--728,
  2019.

\bibitem[Nabian and Meidani(2019)]{Nabian_2019}
Mohammad~Amin Nabian and Hadi Meidani.
\newblock A deep learning solution approach for high-dimensional random
  differential equations.
\newblock \emph{Probabilistic Engineering Mechanics}, 57:\penalty0 14–25, Jul
  2019.
\newblock ISSN 0266-8920.
\newblock \doi{10.1016/j.probengmech.2019.05.001}.
\newblock URL \url{http://dx.doi.org/10.1016/j.probengmech.2019.05.001}.

\bibitem[Raissi et~al.(2017)Raissi, Perdikaris, and
  Karniadakis]{raissi2017physics}
Maziar Raissi, Paris Perdikaris, and George~Em Karniadakis.
\newblock Physics informed deep learning (part i): Data-driven solutions of
  nonlinear partial differential equations.
\newblock \emph{arXiv preprint arXiv:1711.10561}, 2017.

\bibitem[Shi et~al.(2015)Shi, Chen, Wang, and Yeung]{ConvLSTM}
Xingjian Shi, Zhourong Chen, Hao Wang, and Dit-Yan Yeung.
\newblock Convolutional lstm network: A machine learning approach for
  precipitation nowcasting.
\newblock pages 802--810, 2015.

\bibitem[Sun et~al.(2020)Sun, Gao, Pan, and Wang]{sun2020surrogate}
Luning Sun, Han Gao, Shaowu Pan, and Jian-Xun Wang.
\newblock Surrogate modeling for fluid flows based on physics-constrained deep
  learning without simulation data.
\newblock \emph{Computer Methods in Applied Mechanics and Engineering},
  361:\penalty0 112732, 2020.

\bibitem[Yang and Perdikaris(2019)]{yang2019adversarial}
Yibo Yang and Paris Perdikaris.
\newblock Adversarial uncertainty quantification in physics-informed neural
  networks.
\newblock \emph{Journal of Computational Physics}, 394:\penalty0 136--152,
  2019.

\bibitem[Zhu et~al.(2019)Zhu, Zabaras, Koutsourelakis, and
  Perdikaris]{zhu2019physics}
Yinhao Zhu, Nicholas Zabaras, Phaedon-Stelios Koutsourelakis, and Paris
  Perdikaris.
\newblock Physics-constrained deep learning for high-dimensional surrogate
  modeling and uncertainty quantification without labeled data.
\newblock \emph{Journal of Computational Physics}, 394:\penalty0 56--81, 2019.

\end{thebibliography}
}

\end{document}